\pdfoutput=1

\documentclass[11pt]{article}

\usepackage[]{acl}

\usepackage{times}
\usepackage{latexsym}

\usepackage{multirow}
\usepackage{amssymb}
\usepackage{pifont}
\usepackage{url}
\usepackage{graphicx}
\usepackage{booktabs}
\usepackage{appendix}
\usepackage{amsmath}
\usepackage{subcaption}
\usepackage{tcolorbox}
\usepackage{enumitem}

\usepackage[T1]{fontenc}

\usepackage[utf8]{inputenc}

\usepackage{microtype}

\title{Multiple-Choice Question Generation: Towards an Automated Assessment Framework}

\author{Vatsal Raina \\
  Cambridge University  \\
  \texttt{vr311@cam.ac.uk} \\\And
  Mark Gales \\
  Cambridge University \\
  \texttt{mjfg@cam.ac.uk} \\}

\begin{document}
\maketitle

\begin{abstract}

Automated question generation is an important approach to enable personalisation of English comprehension assessment. Recently, transformer-based pretrained language models have demonstrated the ability to produce appropriate  questions from a context paragraph. Typically, these systems are evaluated against a reference set of manually generated questions using n-gram based metrics, or manual qualitative assessment. Here, we focus on a fully automated multiple-choice question generation (MCQG) system where both the question and possible answers must be generated from the context paragraph. Applying n-gram based approaches is challenging for this form of system as the reference set is unlikely to capture the full range of possible questions and answer options. Conversely manual assessment scales poorly and is expensive for MCQG system development. In this work, we propose a set of performance criteria that assess different aspects of the generated multiple-choice questions of interest. These qualities include: grammatical correctness, answerability, diversity and complexity. Initial systems for each of these metrics are  described, and individually evaluated on standard multiple-choice reading comprehension corpora. 

\end{abstract}

\section{Introduction}
\label{sec:intro}

Question generation systems have many applications such as augmenting question-answering datasets to improve machine comprehension models \cite{Duan2017QuestionGF}. Here, question generation is considered as a tool for education. Specifically, it is required to cater to the increased demand of people learning English \cite{Munandar2015HowDE} who require high quality questions to improve their comprehension abilities. It is expensive to employ experts to generate good quality questions for comprehension passages. Therefore, it would be beneficial to develop systems that can automatically generate questions for a comprehension passage to substantially increase the resources available to students. Several bits of work has been done in literature for automatic question generation. For example, \cite{Kriangchaivech2019QuestionGB} investigates question generation as a tool to assist educators in creating quizzes and tests. They further demonstrate that a current failure mode of question generation systems, such as their BERT-based model, is that they tend to create mostly \textit{what} type of questions. Other work looks at the use of a GPT-2 based model \cite{Lopez2020TransformerbasedEQ} or the introduction of special highlight tokens within the passage \cite{Chan2019ARB}.

Any task that requires the development of good quality systems demands an unbiased and reasonable performance metric that can be used to compare and rank various systems. It is possible to demonstrate that the susceptibility of the performance metric with respect to the number of reference samples is heavily dependent on the distribution of the true posterior distribution over all possible output sequences (see Appendix \ref{sec:app_assess}). Greater the entropy in the true posterior distribution, the slower the growth in performance with respect to the number of reference samples. We are interested in assessing the performance of question generation systems. However, question generation is a high entropy sequence-to-sequence task as there are a large number of possible questions for a given input passage. For real question generation datasets \citep{Chen2018LearningQAL, Du2017LearningTA, Zhao2018ParagraphlevelNQ}, it is too expensive to provide a sufficient number of reference samples per example to account for the high entropy of the question generation task. Using a limited number of reference samples, evaluating question generation systems using generic sequence-to-sequence metrics such as BLEU \citep{papineni-etal-2002-bleu}, ROUGE \citep{Lin2004ROUGEAP} and METEOR \citep{banerjee-lavie-2005-meteor} (see Section \ref{sec:assess_frame}) is inappropriate as there is no reason the generated question should match one of the reference questions provided (as there is high entropy in the true posterior distribution).
In development and assessment of current state-of-the-art question generation systems, researchers attempt to bypass the challenge of a large and diverse output space in two respects: \textit{answer-aware} and \textit{sentence-wise} question generation \citep{wang2018answer, Bao2020UniLMv2PL, Dong2019UnifiedLM, Varanasi2020CopyBERTAU, Li2019ImprovingQG, Kim2019ImprovingNQ}. Answer-aware question generation involves passing the answer alongside the context to generate the question. By forcing the answer beforehand, the breadth of acceptable questions is dramatically reduced and hence typical sequence generation metrics such as BLEU, METEOR and ROUGE are acceptable to compare against the small number of references. A similar purpose is achieved with sentence-wise question generation where the length of input passages is limited to consider single sentences at a time. However, the purpose of question generation systems is not to replicate the reference questions for a given input sentence from a passage. As designers of question generation systems we are interested in the quality of the questions generated. We propose the following qualities of question generation systems that we care about:
\begin{itemize}
    \item \textbf{Grammatical fluidity}: The questions must be grammatically fluent.
    \item \textbf{Answerability}: The answer to the question must be present in the context.
    \item \textbf{Diversity}: A diverse range of questions should be possible on a given contextual passage.
    \item \textbf{Complexity}: Questions must be of varying complexity beyond simple answer locators. 
\end{itemize}

The current approach to question generation that uses single sentences from a passage as the input seriously limits the ability to generate complex questions. In order to be able to assess question generation systems using a limited number of references, it should not be necessary to compromise the quality of questions. Therefore, we propose developing a question assessment framework where the framework should be able to give a quantitative score to each of the qualities: grammatical fluidity, answerability, diversity and complexity.

In our work, we focus on multiple-choice question generation (MCQG) for which it is necessary to be able to generate the question, the correct answer option and distractor answer options. In this set-up it becomes increasingly important to shift from n-gram based matching metrics to assessing the qualities of interest. To our knowledge, this is the first attempt to generate complete questions and answer options without explicitly extracting phrases from the context paragraph or using sentence-wise and answer-aware techniques \citep{Ch2020AutomaticMC}, which necessitates sensible automated assessment approaches.

\section{Related Work}

\citet{zheng-etal-2018-multi} mention that neural text generation tasks such as machine translation (MT), summarisation and image captioning often have multiple references available for each sample because each input has multiple acceptable output sequences. Moreover, the references provided are only a tiny subset of the exponentially large space of potential references. The authors find that performance on standard evaluation datasets is improved by using multiple references at training time. They also show that performance can be further boosted by generating pseudo-references from the existing references at training time to get up to 50 references per sample; the true posterior distribution is better sub-sampled in this manner. However, the focus here is on the training process and not on handling the lack of references available in the test set.

\citet{dreyer-marcu-2012-hyter} suggest machine translation is a challenging task because it can be described as a one-to-n mapping, meaning there is ambiguity in the translation. The authors state that they believe that all automatic metrics (before this paper), such as BLEU or NIST, for machine translation fail in appropriateness because these metrics rely on the limited number of human references available. If there was access to \textit{all} possible references for a translation, then the common automatic translation metrics would be reasonable but in reality there are very few references provided. Based on the limitations of other metrics, the paper introduces an annotation tool that efficiently creates an exponential number of correct translations for a given sentence. They then present a new evaluation metric, HYTER, that efficiently uses these large reference networks to compute the performance of the translation generated by an MT system. However, it is not clear how an equivalent annotation tool could be used to generate a list of a large number of reference questions to assess question generation systems.

\citet{Qin2015TrulyEM} propose alternative metrics to BLEU and NIST for machine translation in order to use multiple references more effectively. The main idea is that \citet{Doddington2002AutomaticEO} showed that bag-of-words metrics such as BLEU and NIST will not lead to the best possible results if the number of references is increased. Therefore, the alternative metrics proposed by this paper aim to make better use of the multiple references available at evaluation time. Once again, the alternative metric cannot be easily extended to the question generation task as the metrics of interest in question generation include question complexity, diversity and answerability. It is possible to have reference answerable questions but complex and diverse questions cannot be guaranteed in the labels. Therefore, the metric to be used in question generation should minimally rely on the labelled question and instead judge the quality of a generated question in its own right.

\citet{vu-moschitti-2021-ava} introduce AvA as an automatic evaluation approach for question-answering (QA). They train a system that takes and predicts a set of gold standard references (for question-answering) and then return the accuracy of the system. A transformer-based model is used to encode the references and the predictions and a similarity score is essentially calculated in the encoded space. They claim their system is a lot more appropriate for QA than neural MT metrics such as BLEU. However, the challenge in automated question generation is substantially greater than question-answering because question generation as a task has a lot greater entropy in the posterior distribution and hence the few reference samples poorly capture the posterior distribution and consequently are not fair comparison points for predictions, even in the encoder space.

\citet{dugan2022} investigate answer-agnostic (answer-unaware) question generation. They comment on the inappropriateness of n-gram based assessment as answer-unaware question generation is less restricted and hence less controlled than answer-aware question generation. Consequently, \citet{dugan2022} assess their generated questions in terms of the qualities of relevance, interpretability and acceptability. Currently, these qualities are measured using human markers. We look to extend the assessment of our desired qualities with automated approaches.

\section{Multiple-choice question generation}
\label{sec:mcqg}

\begin{figure}[htbp!]
     \centering
     \includegraphics[width=1.0\columnwidth]{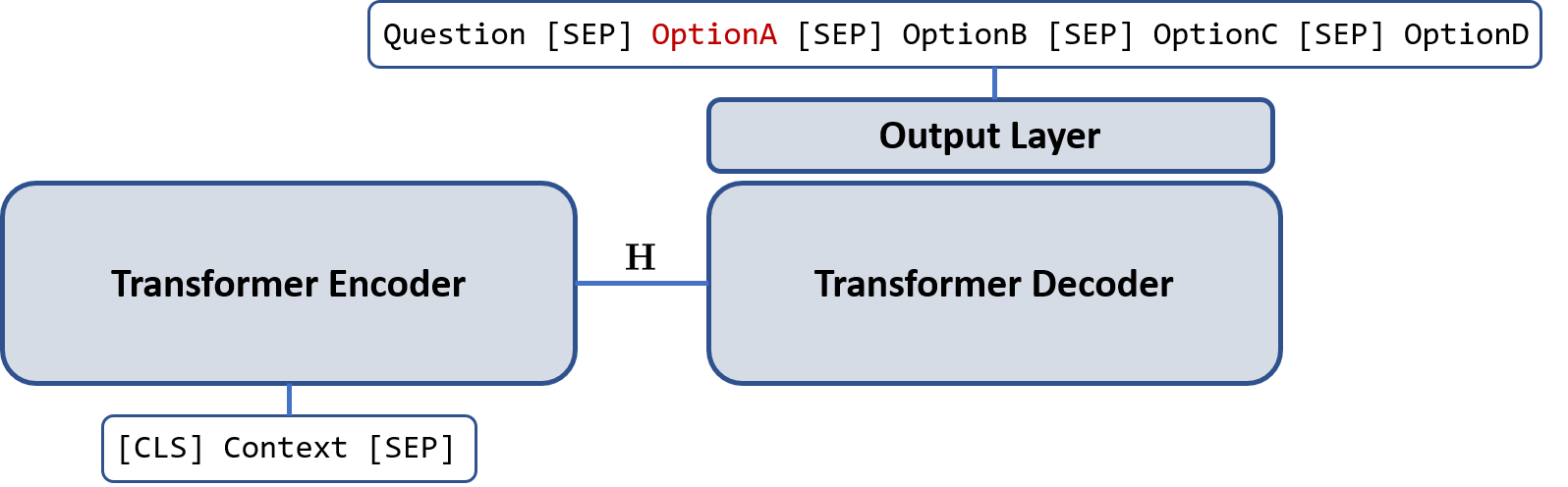}
        \caption{Architecture for multiple-choice question generation.}
        \label{fig:arch}
\end{figure}

Here, we discuss an automated approach to generate multiple-choice questions without constraining the nature of the input context (i.e. maintain answer-unaware question generation). The MCQG task requires a system to generate a question, a correct answer option and distractor answer options when given a context paragraph. State-of-the-art systems for natural language generation predominantly rely on pre-trained language models (PrLMs) \citep{Lewis2020BARTDS, Raffel2020ExploringTL} that are based on the transformer encoder-decoder or just the the transformer decoder architecture \citep{Radford2018ImprovingLU, Radford2019LanguageMA, Brown2020LanguageMA, palm2022lm}. Figure \ref{fig:arch} depicts the vanilla model used in this work for MCQG where the T5 \cite{Raffel2020ExploringTL} PrLM \footnote{Using the \texttt{t5-base} implementation from HuggingFace: \url{https://huggingface.co/t5-base}.} is used for both the transformer encoder and decoder. The generated question and answer options are separated by special \texttt{[SEP]} tokens which allows the question and the answer options to be isolated from one another. The first answer option, \texttt{OptionA}, is treated as the correct answer option while the remaining answer options are treated as distractor answer options.

Additionally, a second baseline is considered using the GPT-3 \citep{Brown2020LanguageMA} model in a zero-shot setting where it is not explicitly trained on the task of question generation. This model is based on the transformer decoder architecture. At inference time, the model takes the context as the input with the following text prepended to the input: \textit{Multiple-choice question with 4 options.}

\section{Assessment Framework}
\label{sec:assess_frame}

Typically, sequence-to-sequence models, including question generation systems, are benchmarked on standard datasets using n-gram based performance metrics that compare the generated sequence against a reference list. However, the task of multiple-choice question generation cannot be assessed appropriately using such n-gram based approaches as the reference list is too narrow to comprehensively cover the possible questions and answer options for a context paragraph \citep{rajpurkar-etal-2016-squad}. Therefore, we propose an assessment framework for the assessment of multiple-choice question generation systems.

Section \ref{sec:mcqg} details the method for the generation of the question and answer options as a single sequence output with each component separated by a special token. Hence, it is necessary to first ensure that four unique answer options are generated. Second, the setup assumes the first answer option generated is the correct answer option to the question. Therefore, it is necessary to ensure a good performing multiple-choice machine reading comprehension (MCMRC) system \citep{Zeng2020ASO} agrees the first answer option is indeed the correct answer option. Next, we describe our fundamental assessment framework based on the qualities of grammatical fluidity, answerability, diversity and complexity (introduced in Section \ref{sec:intro}). As our focus is specifically on multiple-choice question generation, we attach less importance to the criteria of diversity which is more important in other forms of question generation \citep{Du2017LearningTA} and hence we propose a more simplified approach.

\subsection{Grammar}

Grammatical fluidity assesses how well a question abides by the grammatical rules of English language. There are various automated systems available to check for grammatical mistakes in a given sentence such as ERRANT \citep{bryant-etal-2017-automatic}. However, in the realm of pretrained transformed-based language models, grammatical fluidity is of minimal concern as the large amount of training of pretrained language models ensures these systems have an excellent grasp of the English language and consequently make very few grammatical mistakes when used for generating new sentences (including question generation). Here, we quote the grammatical error, $\mathcal{G}$, across all the questions and answer options generated by a question generation system for reference.

\subsection{Answerability}

Answerability is essentially a binary metric that informs whether a given question is answerable based on the provided passage. If the answer to the question can be inferred from the corresponding contextual passage, the question is answerable and otherwise the question is unanswerable \citep{rajpurkar-etal-2018-know}.

Unanswerability is a well-explored area in span-based machine reading comprehension \citep{Zhang2021RetrospectiveRF, rajpurkar-etal-2018-know, Lan2020ALBERTAL}, while limited work has been done for multiple-choice machine reading comprehension \citep{raina-gales-2022-answer}. \citet{raina-gales-2022-answer} specifically explores the situation where unanswerable examples are not present at training time and yet unanswerable questions must be identified during deployment. They suggest that ensemble-based uncertainty measures can be used for identifying unanswerable examples at test time. Therefore, in this work we consider using a set of standard ensemble multiple-choice machine reading comprehension models for measuring the unanswerability of the generated questions from our MCQG system. Specifically, we use the expected entropy, $\mathbb{E}[\mathcal{H}]$, as our choice of uncertainty measure:
\begin{equation}
   \mathbb{E}[\mathcal{H}]  = -\frac{1}{K}\sum_{k=1}^K \sum_y P_{\mathcal{M}_k}(y)\log P_{\mathcal{M}_k}(y)
\end{equation}
where $P_{\mathcal{M}_k}$ denotes the discrete probability distribution using the $k^{\textup{th}}$ machine reading comprehension model member of an ensemble of size $K$ and $y\in \{A,B,C,D\}$ (the possible answer options). We quote the unanswerability score, $\mathcal{A}$, as a constituent of our assessment framework.

\subsection{Diversity}

There is a large amount of literature investigating different categories of question types specifically for reading comprehension \citep{day2005developing}. Here, we describe a potential diversity metric used to judge the diversity of the question types in a set of question-context pairs. For a set of questions, they can be categorised into the following question types: \textit{what}, \textit{who}, \textit{when}, \textit{where}, \textit{why}, \textit{how}, \textit{which}, \textit{yes/no}. For clarity, \textit{yes/no} questions include questions that start with phrases such as \textit{is}, \textit{do} or \textit{does}. A quantitative measure of diversity, $\mathcal{D}$, is proposed as the entropy over the discrete classes:

\begin{equation}
\label{eq:diversity}
    \mathcal{D} = \mathcal{H}(Q) = -\sum_{q \in \mathcal{Q}} q\log q
\end{equation}

where $Q$ denotes the discrete probability distribution over the question types in the set of question-context pairs being considered and $\mathcal{Q}$ is space of all question types. Note, it is fairly trivial to show that the entropy of a discrete probability distribution is a linear transformation of the KL divergence of $Q$ from a uniform distribution. Hence, the diversity metric can be interpreted as a measure of closeness to the uniform distribution over the question type classes. 

In this work, the focus is on multiple-choice reading comprehension questions where the traditional categories of question types may not be the most appropriate. Hence, a simplified number of categories is considered where there are only two categories of questions. The two categories are whether a question makes sense stand-alone or not. For example, \textit{When was King Henry born} makes sense in its own right while \textit{What is the best title for this passage} relies on the options to be read for the question to make sense. The latter type of question can be easily identified in practice with the occurence of the word \textit{passage} within the question. Hence, Equation \ref{eq:diversity} is applied on the binary discrete probability distribution described with bits used as the units to give a diversity score, $\mathcal{D}$.

\subsection{Complexity}

Complexity is a complex quality to assess. An automated metric for complexity must be able to measure the amount of reasoning that is required for a candidate to answer a question in the multiple-choice machine reading comprehension setup. The complexity measure must take the contextual passage into account too as the difficulty of a question is dictated by the choice of words in the associated passage.

\citet{pmlr-v101-liang19a} who introduce the RACE$++$ dataset state that the question complexity in a comprehension exercise is directly correlated with belonging to one of the following question types where these types have been ranked from easiest to the most challenging.
1.\textbf{Word matching}: the question is an extract in the passage and the answer is straightforward.
2. \textbf{Paraphrasing}: the question paraphrases a given sentence from the passage as the answer is an extract of this sentence (SQuAD \citep{rajpurkar-etal-2016-squad} style questions).
3. \textbf{Single-sentence reasoning}: the answer can be deduced from a single sentence of the passage.
4. \textbf{Multi-sentence reasoning}: the answer has to be inferred from connecting information distributed across the passage i.e. over multiple sentences.
5. \textbf{Insufficient/ambiguous}: the question has no answer or the answer is not unique.

\citet{cheng2021guiding} propose difficulty-controllable question generation through step-by-step rewriting. Their proposed model is able to generate questions at required difficulty levels. They motivate the need for difficulty controlled question generation to be used as a tool for curriculum-learning-based methods for QA systems as well as our motivation of designing exams of different difficulty levels for educational purposes. They discuss \citet{ijcai2019-690} as the only other work that defines question difficulty, which in turn defines question difficulty as whether a question-answering model can correctly answer the question. However, it is clear that this definition conflates answerability and difficulty (the terminology complexity is used for difficulty in our work). Hence, the paper redefines the difficulty level of a question as:
\textit{The number of inference steps required to answer it.}
This definition is based upon the number of reasoning hops required to be able to deduce the answer. In order to develop a difficulty-controlled question generation (DCQG) framework, \citet{cheng2021guiding} insist that it is important for the QG model to have a strong grasp over the logic and reasoning complexity of generated questions. Therefore, graph-based methods are the natural choice for such logic modelling. The approach is to convert raw text into a context (entity) graph, from which the answer is sampled as one of the entities. They then design a question generator and question rewriter that initially generates a simple question and then the question rewriter step-by-step converts it into more complex questions by feeding backwards through the reasoning chain. They train using HotpotQA \citep{Yang2018HotpotQAAD} for which all questions require two inference steps which can be decomposed into 1-hop questions. By learning how to convert 1-hop to 2-hop questions, the question rewriter is able to extend the 2-hop questions into n-hop questions and hence generate more complex questions. The paper discusses how to generate complex (difficult) questions but it does not mention how the question complexity can be \textit{assessed} of a given question-passage pair.

In this work, we directly use the RACE$++$ dataset \citep{pmlr-v101-liang19a} to train a deep learning model to explicitly class a multiple-choice question in the complexity levels of easy, medium and hard. The RACE$++$ dataset is partitioned into easy, medium and hard questions (see Section \ref{sec:data}) as annotated by human examiners. Therefore, a system can be trained to classify a given question as either easy, medium or hard. Typically, deep learning models output a probability distribution over the three classes such that $p_{\text{easy}} + p_{\text{medium}} + p_{\text{hard}} = 1$. The complexity, $\mathcal{C}$, of a given question is then calculated as $0.0\times p_{\text{easy}} + 0.5\times p_{\text{medium}} + 1.0\times p_{\text{hard}} $. As a part of our assessment framework for multiple-choice question generation, we report the mean $\mathcal{C}$
across an evaluation set.


\section{Experiments}

\subsection{Data}
\label{sec:data}

Our experiments are primarily based on the multiple-choice machine reading comprehension dataset RACE++ \citep{pmlr-v101-liang19a} which requires a candidate to select the correct answer option from a possible choice of 4. RACE++ extends a standard benchmarking dataset, RACE \citep{Lai2017RACELR} for MCMRC. RACE++ is comprised of English comprehension questions from middle school (RACE-M), high school (RACE-H) and college level (RACE-C). Hence, RACE-M, RACE-H and RACE-C can respectively be treated as easy, medium and hard questions. Table \ref{tab:race} details the number of questions in each of these subsets for the training (Trn), development (Dev) and evaluation (Evl) splits. In particular, the Trn split consists of 100,568 questions with 25.3/62.1/12.6 \% respectively for RACE-M, RACE-H and RACE-C. Similarly, Dev and Evl splits are dominated with questions from RACE-H with a total number of questions of 5,599 and 5,642 respectively.

\begin{table}[htbp!]
\centering
\begin{small}
    \begin{tabular}{ll|rrr}
    \toprule
Dataset & & Trn & Dev & Evl \\
\midrule
\multirow{2}*{RACE-M}
& Questions & 25,421 & 1,436 & 1,436 \\
& Contexts & 6,409 & 368 & 362 \\
\midrule
\multirow{2}*{RACE-H}
& Questions & 62,445 & 3,451 & 3,498 \\
& Contexts & 18,728 & 1,021 & 1,045 \\
\midrule
\multirow{2}*{RACE-C}
& Questions & 12,702 & 712 & 708 \\
& Contexts & 2,437 & 136 & 135 \\
   \bottomrule
    \end{tabular}
    \end{small}
\caption{RACE++ data statistics.}
    \label{tab:race}
\end{table}

\subsection{Setup}
\label{sec:setup}

This section describes the various experiments in order to be able fulfill the criteria detailed in the assessment framework of Section \ref{sec:assess_frame}. All models have hyperparamter tuning performed on the Dev split of RACE++. 

First an ensemble of 3 models is trained to perform the vanilla MCMRC task on the RACE++ dataset. The architecture of the model is based on the baseline systems from \citet{Yu2020ReClorAR} with the ELECTRA PrLM \cite{clark2020electra} specifically selected based on the high performance demonstrated in MCMRC tasks \citep{raina-gales-2022-answer}. The input to the transformer model is constructed as
\begin{scriptsize}
\texttt{[CLS] Context [SEP] Question Option [SEP] [PAD] ...}
\end{scriptsize}
where the context followed by the question concatenated with an option are separated by a special [SEP] token. The construct is repeated for each of the answer options such that the four sequences are inputted in parallel to the transformer encoder architecture (weights shared for each of the four sequences) that is followed by a classification head to return a probability distribution over the answer options. At inference, the predicted answer option is selected to be the one with the greatest probability mass associated with it.
See Appendix \ref{sec:app_trn_mcmrc} for details about hyperparameter tuning.

Second, an ensemble of 3 models are trained to determine the question complexity of a system. These models are trained on the RACE$++$ splits but the classification labels used here are either easy, medium or hard corresponding to RACE-M, RACE-H and RACE-C respectively. Similar, to the MCMRC system, an ELECTRA-based model \footnote{\texttt{ELECTRA-large} from \url{https://huggingface.co/google/electra-large-discriminator}.} with a three-way classification head is used but the input to this model is of the form 
\begin{scriptsize}
\texttt{[CLS] Question [SEP] Context [SEP] OptionA [SEP] OptionB [SEP] OptionC [SEP] OptionD [SEP] [PAD] ...}
\end{scriptsize}
\footnote{Several other architectures were considered but the proposed one is the simplest and best performing.}
where the question, context and all four answer options are concatenated together with the special [SEP] tokens. See Appendix \ref{sec:app_trn_qc} for further training details and hyperparamter tuning.

A single question generation system based on the architecture of Figure \ref{fig:arch} from Section \ref{sec:mcqg} is trained with teacher forcing \citep{NIPS2016_16026d60} based on the T5 PrLM \citep{Raffel2020ExploringTL}. At inference time, deterministic beam search with a beam size of 4 \footnote{Various stochastic decoding approaches were considered including top-K-sampling and top-p-sampling but it was found the deterministic beam search generated the best quality questions.} is used with a single question and set of answer options generated on each context from RACE++. Further details with regard to hyperparameter tuning are given in Appendix \ref{sec:app_trn_qg}. A second question generation model is trained using GPT-3 in the zero-shot setting \footnote{Latest model provided by OpenAI: text-davinci-002 in completion mode where at generation time the temperature is set to 0.4, maximum length to 1024 tokens, top P to 1 and the frequency and presence penalties to 0.}.

The qualities described in the assessment framework \footnote{Code provided if accepted.} of Section \ref{sec:assess_frame} are analysed on the questions generated from the T5 system. Particularly, the ELECTRA-based MCMRC system is used to derive the uncertainty scores and the ELECTRA-based question complexity (QC) system is used to derive the complexity scores. As the question generation system is uncontrolled, it is additionally important to assess what fraction of generated outputs actually have the desired number of 4 unique answer options. It is also necessary to assess the \textit{accuracy} of the generated questions. From Section \ref{sec:mcqg} it is expected that the first answer option generated should be the correct answer option to the generated question. Therefore, the accuracy is assessed as the fraction of generated samples where each of the three MCMRC systems in the ensemble agree at inference time that the first answer options is most likely the correct answer.


\section{Results}

This section presents the main results of the paper. First, the baseline performance of the vanilla systems for MCMRC and QC classification are given. Then the results of the MCQG system are given using the proposed assessment framework based on the MCMRC and QC systems.

\subsection{Machine reading comprehension}

\begin{table}[htbp!]
\centering
\begin{small}
    \begin{tabular}{l|rrrr}
    \toprule
System & M & H & C & All \\
\midrule
ELECTRA, single & 88.30 & 84.11 & 81.07 & 84.80  \\
ELECTRA, ensemble & 88.09 & 84.42 & 81.64 & 85.01 \\
   \bottomrule
    \end{tabular}
    \end{small}
\caption{Multiple-choice machine reading comprehension on Evl of RACE++ assessed using accuracy (\%). See Appendix \ref{sec:app_additional_mcmrc} for standard deviations of single models.}
    \label{tab:res_mrc}
\end{table}

Table \ref{tab:res_mrc} presents the performance of the ELECTRA-based system on the RACE++ dataset \footnote{\citet{pmlr-v101-liang19a} present baseline models on RACE++ using BERT-base but these results are omitted here for comparison due to their significantly lower performance compared to ELECTRA-large.} for multiple-choice machine reading comprehension. It is observed that the ensembling of the ELECTRA models boosts the overall performance from an accuracy of 84.80\% to 85.01\% on the evaluation split. More fine-grained results are also provided on the RACE-M, RACE-H and RACE-C splits of the dataset. As expected, the systems struggle the most on the college level questions and find the middle school level questions the easiest with both the single and ensembled systems achieving approximately 7\% greater accuracy on RACE-M compared to RACE-C. From these results that the overall performance of greater than 80\%, it is somewhat reasonable to trust these MCMRC models as potential systems to be used as part of the MCQG assessment process in the proposed framework.

\subsection{Question complexity}

\begin{table}[htbp!]
\centering
\begin{small}
    \begin{tabular}{l|rr|rr}
    \toprule
\multirow{2}{*}{System}  & \multicolumn{2}{c|}{\texttt{Accuracy}} &  \multicolumn{2}{c}{\texttt{Macro F1}} \\
 & Dev & Evl & Dev & Evl \\
\midrule
Majority-class & 61.64 & 62.00 & 25.42 & 25.51 \\
Vocab-based & 63.87 & 63.98 & 34.31 & 33.84 \\
ELECTRA, single & 86.00 & 86.19 & 83.92 & 84.66 \\
ELECTRA, ensemble & \textbf{86.66} & \textbf{86.97} & \textbf{84.65} & \textbf{85.52} \\
   \bottomrule
    \end{tabular}
    \end{small}
\caption{Question complexity results. See Appendix \ref{sec:app_additional_qc} for standard deviations of single-seed results.}
    \label{tab:res_comp}
\end{table}

Table \ref{tab:res_comp} presents the results of various QC systems in terms of accuracy and macro F1 \citep{yang1999re} for classification between the easy (RACE-M), medium (RACE-H) and hard (RACE-C) question classes. As a baseline, the majority-class system is considered that always predicts medium as the complexity class (RACE-H is the dominant subset from Table \ref{tab:race}). Next, a second baseline as a structured vocabulary (vocab) based system is considered. Here, standard vocabulary lists \footnote{These lists are based on US English. Source not given to preserve anonymity.} exist that have partitioned the whole vocabulary into the categories of beginner, intermediate and expert. Hence, each question, context and set of answer options has a structured complexity score calculated by determining the average vocabulary score for each example where beginner words have a score of 0.0, intermediate of 0.5 and expert of 1.0. This structured vocab-based score (with the optimal complexity thresholds between easy/medium and medium/hard determined on the Dev split) observes a performance boost compared to the majority-class by about 2\% accuracy and 9\% macro F1. Finally, the results of the single and ensemble versions of the ELECTRA model described in Section \ref{sec:setup} are presented. Significantly better performance is observed with the ensembled system achieving an accuracy of 86.97\% and macro F1 of 85.52\% on the Evl split. Therefore, this QC is considered to be sufficiently well performing to use as a part of the MCQG assessment framework.
\begin{figure}[htbp!]
     \centering
     \includegraphics[width=1.0\columnwidth]{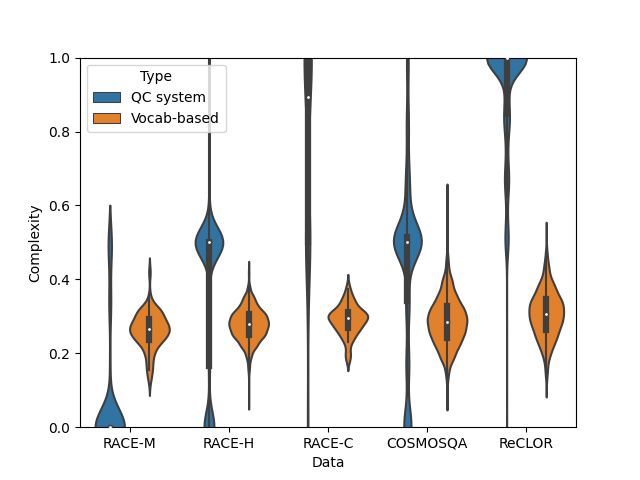}
        \caption{Complexity scores calculated using proposed QC system and structured vocabulary-based scores.}
        \label{fig:boxplot}
\end{figure}

Figure \ref{fig:boxplot} depicts the range of complexity scores on the evaluation splits of various standard MCMRC datasets using the ensembled ELECTRA model (QC system) and the strucutured vocab-based system. The three parts of RACE++ (RACE-M, RACE-H and RACE-C) are shown as well as two additional datasets COSMOSQA \citep{Huang2019CosmosQM} and ReCLOR \citep{Yu2020ReClorAR}. It is seen the complexity scores assigned to RACE-M, RACE-H and RACE-C are distinct with minimal overlap using the QC system while the vocab-based system observes a more gradual increase. Applying these complexity systems to COSMOSQA and ReCLOR, it is reassuring to see that ReCLOR has the greatest complexity scores given it is observed to have the most complex multiple-choice questions \citep{raina-gales-2022-answer} requiring logical reasoning.  COSMOSQA has a complexity similar to RACE-H.

\subsection{Question generation}

In our results here, we present the assessment of our MCQG system (described in Section \ref{sec:mcqg}) using our proposed framework. In particular, we assess the grammatical fluidity ($\mathcal{G}$), unanswerability ($\mathcal{A}$), diversity ($\mathcal{D}$) and the complexity ($\mathcal{C}$) of the generated questions with the corresponding answer options. Additionally, it is necessary to assess what fraction of the questions actually have 4 unique options (4 opt) as the MCQG model does not guarantee that this will be the case. The accuracy (acc) is also reported (see Section \ref{sec:assess_frame}) as an additional check to confirm what fraction of the generated questions and answer option sets do actually ensure the first answer option is the correct one. 

Table \ref{tab:res_mcqg} first presents the results of the assessment framework on the default human generated questions of the RACE++ dataset for the Evl split in order to provide an upperbound of performance for MCQG system (T5, all) where one question with the corresponding answer options has been generated per context. A filtered set of generated questions (T5, filt) is considered too where each of the questions has 4 unqiue answer options and the first answer option has been established to be the correct one. This filtered set consists of a set of 354 questions from an initial total of 1,542 (Table \ref{tab:race}).
For comparison, the results for a baseline based on GPT-3 zero-shot (GPT-3) are presented too. Note, the accuracy of the GPT-3 model is not given as it does not inform which of the answer options it generates is the correct answer.
In terms of grammatical fluidity, pretrained language models have an excellent grasp of the rules of English language and hence such models generate grammatically correct questions and answer options, which leads to both the T5 and GPT-3 systems to achieve no grammatical errors (to 4 significant figures) according to the ERRANT system, reinforcing that state-of-the-art generation systems no longer need to be concerned in making mistakes with regard to grammatical structure.
The unanswerability rate of the T5 generated questions is substantially higher than the human based system but is reduced by about 25\% when considering the filtered set of questions. The GPT-3 system tends to naturally generate more answerable questions than the T5 model. In terms of complexity, although the MCQG system is trained using questions of mean complexity of 0.4402, the generated questions tend to be easier in nature with a slight further reduction in the complexity when considering the filtered set. The complexity of GPT-3 appears to be comparable to the T5 system.

\begin{table}[htbp!]
\centering
\begin{small}
    \begin{tabular}{l|cc|cccc}
    \toprule
    System & 4 opts & Acc  & $\mathcal{A} (\downarrow)$ & $\mathcal{C} (\uparrow)$ & $\mathcal{D} (\uparrow)$ \\
    \midrule
Human & 100 & 86.97 & 0.2740 & 0.4402 & 0.7750 \\
T5, all & 77.24 & 43.77  & 0.8413 & 0.3950 & 0.6222 \\
GPT-3 & 80.93 & -- & 0.7110 & 0.4099 & 0.5104 \\
\midrule
T5, filt & 100 & 100 & 0.6350 & 0.3839 & 0.6629 \\

   \bottomrule
    \end{tabular}
    \end{small}
\caption{Question generation assessment on Evl with un$\mathcal{A}$nswerability, $\mathcal{C}$omplexity and $\mathcal{D}$iversity.}
    \label{tab:res_mcqg}
\end{table}

Additionally, the quality of the T5 question generation system is further assessed by investigating its effectiveness for augmenting the training data for multiple-choice machine reading comprehension. The MCQG system is used to generate questions on the contexts from Trn. These questions are filtered as described above to remove questions that do not 
have 4 unique options or have any disagreement in predictions between the baseline MCMRC systems in the ensemble, leaving 5,766 questions. The set of filtered questions are used as extra augmentation data for the training of an ensembled MCMRC system. The augmented ensembled system observes a marginal performance boost on the MCMRC task for RACE$++$ of 0.36\% on Dev and 0.12\% on Evl. Future work can look at more effective filtration strategies for greater gains in the question generation augmented MCMRC task. 

As a limitation of the current approach for MCQG \footnote{Full list of generated questions can be reproduced with described training and inference process.}, it is observed that some of the generated distractor options are in fact valid answer options (see Appendix \ref{sec:app_example}). Further work should investigate explicit methods of ensuring the distractor options do not answer the question. 

\section{Conclusions}

This work aims to propose a sensible assessment framework for multiple choice question generation in order to encourage the question generation community to consider more appropriate methods of assessing and benchmarking developed systems to reflect the qualities of interest rather than arbitrary n-gram based approaches. Here, the first fully automated end-to-end multiple-choice question generation system is proposed for generating a question, the correct answer and distractor options for an input context without relying on explicit phrase extraction based techniques. 

To assess the quality of the generated questions sensibly, an assessment framework is proposed that rates the grammatical fluidity, answerability, diversity and complexity of the questions. Standard tools are considered for assessing the rate of grammatical errors. The unanswerability of the questions is assessed by using the predictive uncertainty estimates in the predictions of an ensemble of high-performing multiple-choice reading comprehension systems. Diversity is measured as the entropy over question types. Finally, complexity assessment as a single score is captured by a directly trained supervised system to perform three-way classification between easy, medium and hard questions.

The automated assessment of high-entropy sequence-to-sequence models is critical for accelerating progress in many natural language generation tasks, and hence it is important that future work continues to consider and refine appropriate assessment frameworks to comprehensively understand the generation quality of novel systems.

\section{Limitations}

Here, the limitations of the current approaches are discussed. First, both measures of unanswerability and question complexity are model and corpus-specific. Hence, it is not clear about the applicability of these metrics beyond the RACE++ dataset. Specifically, the unanswerability measure runs the risk of conflating answerability with the failure of a reading comprehension question. The validity of such metrics is centred on in-domain data, which may be diminishingly effective at discriminating the performance of generative models on shifted data (e.g the nature of questions in ReClor is more logical-based and requires higher inference than questions in RACE++, which might make measures trained on RACE++ struggle at assessment on questions generated in the ReClor style). No human evaluation is performed of whether the assessment metrics correlate explicitly with human notions of answerability and complexity. Hence, further work should invest resources to comprehensively establish the validity of the proposed measures. Finally, the question complexity system is trained specifically on the meaning of complexity as described for RACE++. However, question complexity has several definitions and hence further work should establish whether this interpretation of complexity may align with other views of complexity.

\bibliography{anthology,custom}

\begin{thebibliography}{48}
\expandafter\ifx\csname natexlab\endcsname\relax\def\natexlab#1{#1}\fi

\bibitem[{Banerjee and Lavie(2005)}]{banerjee-lavie-2005-meteor}
Satanjeev Banerjee and Alon Lavie. 2005.
\newblock \href {https://www.aclweb.org/anthology/W05-0909} {{METEOR}: An
  automatic metric for {MT} evaluation with improved correlation with human
  judgments}.
\newblock In \emph{Proceedings of the {ACL} Workshop on Intrinsic and Extrinsic
  Evaluation Measures for Machine Translation and/or Summarization}, pages
  65--72, Ann Arbor, Michigan. Association for Computational Linguistics.

\bibitem[{Bao et~al.(2020)Bao, Dong, Wei, Wang, Yang, Liu, Wang, Piao, Gao,
  Zhou et~al.}]{Bao2020UniLMv2PL}
Hangbo Bao, Li~Dong, Furu Wei, Wenhui Wang, Nan Yang, Xiaodong Liu, Yu~Wang,
  Songhao Piao, Jianfeng Gao, Ming Zhou, et~al. 2020.
\newblock Unilmv2: pseudo-masked language models for unified language model
  pre-training.
\newblock pages 642--652.

\bibitem[{Brown et~al.(2020)Brown, Mann, Ryder, Subbiah, Kaplan, Dhariwal,
  Neelakantan, Shyam, Sastry, Askell, Agarwal, Herbert-Voss, Krueger, Henighan,
  Child, Ramesh, Ziegler, Wu, Winter, Hesse, Chen, Sigler, Litwin, Gray, Chess,
  Clark, Berner, McCandlish, Radford, Sutskever, and
  Amodei}]{Brown2020LanguageMA}
Tom~B. Brown, Benjamin Mann, Nick Ryder, Melanie Subbiah, J.~Kaplan, Prafulla
  Dhariwal, Arvind Neelakantan, Pranav Shyam, Girish Sastry, Amanda Askell,
  Sandhini Agarwal, Ariel Herbert-Voss, Gretchen Krueger, T.~Henighan,
  R.~Child, A.~Ramesh, Daniel~M. Ziegler, Jeff Wu, Clemens Winter, Christopher
  Hesse, Mark Chen, Eric Sigler, Mateusz Litwin, Scott Gray, Benjamin Chess,
  Jack Clark, Christopher Berner, Sam McCandlish, Alec Radford, Ilya Sutskever,
  and Dario Amodei. 2020.
\newblock Language models are few-shot learners.
\newblock \emph{Advances in neural information processing systems},
  33:1877--1901.

\bibitem[{Bryant et~al.(2017)}]{bryant-etal-2017-automatic}
Christopher Bryant et~al. 2017.
\newblock \href {https://doi.org/10.18653/v1/P17-1074} {Automatic annotation
  and evaluation of error types for grammatical error correction}.
\newblock In \emph{Proceedings of the 55th Annual Meeting of the Association
  for Computational Linguistics (Volume 1: Long Papers)}, pages 793--805,
  Vancouver, Canada. Association for Computational Linguistics.

\bibitem[{Ch and Saha(2020)}]{Ch2020AutomaticMC}
Dhawaleswar~Rao Ch and Sujan~Kumar Saha. 2020.
\newblock Automatic multiple choice question generation from text: A survey.
\newblock \emph{IEEE Transactions on Learning Technologies}, 13:14--25.

\bibitem[{Chan and Fan(2019)}]{Chan2019ARB}
Ying-Hong Chan and Yao-Chung Fan. 2019.
\newblock A recurrent bert-based model for question generation.
\newblock In \emph{Proceedings of the 2nd Workshop on Machine Reading for
  Question Answering}, pages 154--162.

\bibitem[{Chen et~al.(2018)Chen, Yang, Hauff, and Houben}]{Chen2018LearningQAL}
Guanliang Chen, Jie Yang, Claudia Hauff, and Geert-Jan Houben. 2018.
\newblock Learningq: a large-scale dataset for educational question generation.
\newblock In \emph{Twelfth International AAAI Conference on Web and Social
  Media}.

\bibitem[{Cheng et~al.(2021)Cheng, Li, Liu, Zhao, Li, Lin, and
  Zheng}]{cheng2021guiding}
Yi~Cheng, Siyao Li, Bang Liu, Ruihui Zhao, Sujian Li, Chenghua Lin, and Yefeng
  Zheng. 2021.
\newblock Guiding the growth: Difficulty-controllable question generation
  through step-by-step rewriting.

\bibitem[{Chowdhery et~al.(2022)Chowdhery, Narang, Devlin, Bosma, Mishra,
  Roberts, Barham, Chung, Sutton, Gehrmann, Schuh, Shi, Tsvyashchenko, Maynez,
  Rao, Barnes, Tay, Shazeer, Prabhakaran, Reif, Du, Hutchinson, Pope, Bradbury,
  Austin, Isard, Gur-Ari, Yin, Duke, Levskaya, Ghemawat, Dev, Michalewski,
  Garcia, Misra, Robinson, Fedus, Zhou, Ippolito, Luan, Lim, Zoph, Spiridonov,
  Sepassi, Dohan, Agrawal, Omernick, Dai, Pillai, Pellat, Lewkowycz, Moreira,
  Child, Polozov, Lee, Zhou, Wang, Saeta, Diaz, Firat, Catasta, Wei,
  Meier-Hellstern, Eck, Dean, Petrov, and Fiedel}]{palm2022lm}
Aakanksha Chowdhery, Sharan Narang, Jacob Devlin, Maarten Bosma, Gaurav Mishra,
  Adam Roberts, Paul Barham, Hyung~Won Chung, Charles Sutton, Sebastian
  Gehrmann, Parker Schuh, Kensen Shi, Sasha Tsvyashchenko, Joshua Maynez,
  Abhishek Rao, Parker Barnes, Yi~Tay, Noam Shazeer, Vinodkumar Prabhakaran,
  Emily Reif, Nan Du, Ben Hutchinson, Reiner Pope, James Bradbury, Jacob
  Austin, Michael Isard, Guy Gur-Ari, Pengcheng Yin, Toju Duke, Anselm
  Levskaya, Sanjay Ghemawat, Sunipa Dev, Henryk Michalewski, Xavier Garcia,
  Vedant Misra, Kevin Robinson, Liam Fedus, Denny Zhou, Daphne Ippolito, David
  Luan, Hyeontaek Lim, Barret Zoph, Alexander Spiridonov, Ryan Sepassi, David
  Dohan, Shivani Agrawal, Mark Omernick, Andrew~M. Dai,
  Thanumalayan~Sankaranarayana Pillai, Marie Pellat, Aitor Lewkowycz, Erica
  Moreira, Rewon Child, Oleksandr Polozov, Katherine Lee, Zongwei Zhou, Xuezhi
  Wang, Brennan Saeta, Mark Diaz, Orhan Firat, Michele Catasta, Jason Wei,
  Kathy Meier-Hellstern, Douglas Eck, Jeff Dean, Slav Petrov, and Noah Fiedel.
  2022.
\newblock \href {https://doi.org/10.48550/ARXIV.2204.02311} {Palm: Scaling
  language modeling with pathways}.

\bibitem[{Clark et~al.(2020)Clark, Luong, Le, and Manning}]{clark2020electra}
Kevin Clark, Minh-Thang Luong, Quoc~V. Le, and Christopher~D. Manning. 2020.
\newblock Electra: Pre-training text encoders as discriminators rather than
  generators.
\newblock \emph{International Conference on Learning Representations},
  abs/2003.10555.

\bibitem[{Day and Park(2005)}]{day2005developing}
R.~Day and J.-s Park. 2005.
\newblock Developing reading comprehension questions.
\newblock \emph{Reading in a Foreign Language}, 17:60--73.

\bibitem[{Doddington(2002)}]{Doddington2002AutomaticEO}
George Doddington. 2002.
\newblock Automatic evaluation of machine translation quality using n-gram
  co-occurrence statistics.
\newblock In \emph{Proceedings of the second international conference on Human
  Language Technology Research}, pages 138--145.

\bibitem[{Dong et~al.(2019)Dong, Yang, Wang, Wei, Liu, Wang, Gao, Zhou, and
  Hon}]{Dong2019UnifiedLM}
Li~Dong, Nan Yang, Wenhui Wang, Furu Wei, Xiaodong Liu, Yu~Wang, Jianfeng Gao,
  Ming Zhou, and Hsiao-Wuen Hon. 2019.
\newblock Unified language model pre-training for natural language
  understanding and generation.
\newblock \emph{Advances in Neural Information Processing Systems}, 32.

\bibitem[{Dreyer and Marcu(2012)}]{dreyer-marcu-2012-hyter}
Markus Dreyer and Daniel Marcu. 2012.
\newblock \href {https://www.aclweb.org/anthology/N12-1017} {{H}y{TER}:
  Meaning-equivalent semantics for translation evaluation}.
\newblock In \emph{Proceedings of the 2012 Conference of the North {A}merican
  Chapter of the Association for Computational Linguistics: Human Language
  Technologies}, pages 162--171, Montr{\'e}al, Canada. Association for
  Computational Linguistics.

\bibitem[{Du et~al.(2017)Du, Shao, and Cardie}]{Du2017LearningTA}
Xinya Du, Junru Shao, and Claire Cardie. 2017.
\newblock Learning to ask: Neural question generation for reading
  comprehension.
\newblock pages 1342--1352.

\bibitem[{Duan et~al.(2017)Duan, Tang, Chen, and Zhou}]{Duan2017QuestionGF}
Nan Duan, Duyu Tang, Peng Chen, and Ming Zhou. 2017.
\newblock Question generation for question answering.
\newblock In \emph{Proceedings of the 2017 conference on empirical methods in
  natural language processing}, pages 866--874.

\bibitem[{Dugan et~al.(2022)Dugan, Miltsakaki, Upadhyay, Ginsberg, Gonzalez,
  Choi, Yuan, and Callison-Burch}]{dugan2022}
Liam Dugan, Eleni Miltsakaki, Shriyash Upadhyay, Etan Ginsberg, Hannah
  Gonzalez, DaHyeon Choi, Chuning Yuan, and Chris Callison-Burch. 2022.
\newblock \href {https://doi.org/10.18653/v1/2022.findings-acl.151} {A
  feasibility study of answer-agnostic question generation for education}.
\newblock In \emph{Findings of the Association for Computational Linguistics:
  ACL 2022}, pages 1919--1926, Dublin, Ireland. Association for Computational
  Linguistics.

\bibitem[{Gao et~al.(2019)Gao, Bing, Chen, Lyu, and King}]{ijcai2019-690}
Yifan Gao, Lidong Bing, Wang Chen, Michael Lyu, and Irwin King. 2019.
\newblock \href {https://doi.org/10.24963/ijcai.2019/690} {Difficulty
  controllable generation of reading comprehension questions}.
\newblock In \emph{Proceedings of the Twenty-Eighth International Joint
  Conference on Artificial Intelligence, {IJCAI-19}}, pages 4968--4974.
  International Joint Conferences on Artificial Intelligence Organization.

\bibitem[{Huang et~al.(2019)Huang, Le~Bras, Bhagavatula, and
  Choi}]{Huang2019CosmosQM}
Lifu Huang, Ronan Le~Bras, Chandra Bhagavatula, and Yejin Choi. 2019.
\newblock \href {https://doi.org/10.18653/v1/D19-1243} {Cosmos {QA}: Machine
  reading comprehension with contextual commonsense reasoning}.
\newblock pages 2391--2401.

\bibitem[{Kim et~al.(2019)Kim, Lee, Shin, and Jung}]{Kim2019ImprovingNQ}
Yanghoon Kim, Hwanhee Lee, Joongbo Shin, and Kyomin Jung. 2019.
\newblock \href {https://doi.org/10.1609/aaai.v33i01.33016602} {Improving
  neural question generation using answer separation}.
\newblock \emph{Proceedings of the AAAI Conference on Artificial Intelligence},
  33:6602--6609.

\bibitem[{Kriangchaivech and Wangperawong(2019)}]{Kriangchaivech2019QuestionGB}
Kettip Kriangchaivech and Artit Wangperawong. 2019.
\newblock Question generation by transformers.
\newblock \emph{arXiv preprint arXiv:1909.05017}.

\bibitem[{Lai et~al.(2017)Lai, Xie, Liu, Yang, and Hovy}]{Lai2017RACELR}
Guokun Lai, Qizhe Xie, Hanxiao Liu, Yiming Yang, and E.~Hovy. 2017.
\newblock Race: Large-scale reading comprehension dataset from examinations.
\newblock In \emph{Proceedings of the 2017 Conference on Empirical Methods in
  Natural Language Processing}.

\bibitem[{Lamb et~al.(2016)Lamb, ALIAS PARTH~GOYAL, Zhang, Zhang, Courville,
  and Bengio}]{NIPS2016_16026d60}
Alex~M Lamb, Anirudh~Goyal ALIAS PARTH~GOYAL, Ying Zhang, Saizheng Zhang,
  Aaron~C Courville, and Yoshua Bengio. 2016.
\newblock \href
  {https://proceedings.neurips.cc/paper/2016/file/16026d60ff9b54410b3435b403afd226-Paper.pdf}
  {Professor forcing: A new algorithm for training recurrent networks}.
\newblock In \emph{Advances in Neural Information Processing Systems},
  volume~29. Curran Associates, Inc.

\bibitem[{Lan et~al.(2019)Lan, Chen, Goodman, Gimpel, Sharma, and
  Soricut}]{Lan2020ALBERTAL}
Zhenzhong Lan, Mingda Chen, Sebastian Goodman, Kevin Gimpel, Piyush Sharma, and
  Radu Soricut. 2019.
\newblock Albert: A lite bert for self-supervised learning of language
  representations.

\bibitem[{Lewis et~al.(2020)Lewis, Liu, Goyal, Ghazvininejad, Mohamed, Levy,
  Stoyanov, and Zettlemoyer}]{Lewis2020BARTDS}
Mike Lewis, Yinhan Liu, Naman Goyal, Marjan Ghazvininejad, Abdelrahman Mohamed,
  Omer Levy, Veselin Stoyanov, and Luke Zettlemoyer. 2020.
\newblock \href {https://doi.org/10.18653/v1/2020.acl-main.703} {{BART}:
  Denoising sequence-to-sequence pre-training for natural language generation,
  translation, and comprehension}.
\newblock In \emph{Proceedings of the 58th Annual Meeting of the Association
  for Computational Linguistics}, pages 7871--7880, Online. Association for
  Computational Linguistics.

\bibitem[{Li et~al.(2019)Li, Gao, Bing, King, and Lyu}]{Li2019ImprovingQG}
Jingjing Li, Yifan Gao, Lidong Bing, Irwin King, and Michael~R Lyu. 2019.
\newblock Improving question generation with to the point context.
\newblock pages 3216--3226.

\bibitem[{Liang et~al.(2019)Liang, Li, and Yin}]{pmlr-v101-liang19a}
Yichan Liang, Jianheng Li, and Jian Yin. 2019.
\newblock \href {http://proceedings.mlr.press/v101/liang19a.html} {A new
  multi-choice reading comprehension dataset for curriculum learning}.
\newblock In \emph{Proceedings of The Eleventh Asian Conference on Machine
  Learning}, volume 101 of \emph{Proceedings of Machine Learning Research},
  pages 742--757, Nagoya, Japan. PMLR.

\bibitem[{Lin(2004)}]{Lin2004ROUGEAP}
Chin-Yew Lin. 2004.
\newblock Rouge: A package for automatic evaluation of summaries.
\newblock In \emph{Text summarization branches out}, pages 74--81.

\bibitem[{Lopez et~al.(2020)Lopez, Cruz, Cruz, and
  Cheng}]{Lopez2020TransformerbasedEQ}
Luis~Enrico Lopez, Diane~Kathryn Cruz, Jan Christian~Blaise Cruz, and Charibeth
  Cheng. 2020.
\newblock Transformer-based end-to-end question generation.
\newblock \emph{arXiv preprint arXiv:2005.01107}, 4.

\bibitem[{Munandar(2015)}]{Munandar2015HowDE}
Imam Munandar. 2015.
\newblock How does english language learning contribute to social mobility of
  language learners?
\newblock \emph{Al-Ta Lim Journal}, 22(3):236--242.

\bibitem[{Papineni et~al.(2002)Papineni, Roukos, Ward, and
  Zhu}]{papineni-etal-2002-bleu}
Kishore Papineni, Salim Roukos, Todd Ward, and Wei-Jing Zhu. 2002.
\newblock \href {https://doi.org/10.3115/1073083.1073135} {{B}leu: a method for
  automatic evaluation of machine translation}.
\newblock In \emph{Proceedings of the 40th Annual Meeting of the Association
  for Computational Linguistics}, pages 311--318, Philadelphia, Pennsylvania,
  USA. Association for Computational Linguistics.

\bibitem[{Qin and Specia(2015)}]{Qin2015TrulyEM}
Ying Qin and Lucia Specia. 2015.
\newblock Truly exploring multiple references for machine translation
  evaluation.
\newblock In \emph{Proceedings of the 18th Annual Conference of the European
  Association for Machine Translation}.

\bibitem[{Radford and Narasimhan(2018)}]{Radford2018ImprovingLU}
Alec Radford and Karthik Narasimhan. 2018.
\newblock Improving language understanding by generative pre-training.

\bibitem[{Radford et~al.(2019)Radford, Wu, Child, Luan, Amodei, Sutskever
  et~al.}]{Radford2019LanguageMA}
Alec Radford, Jeffrey Wu, Rewon Child, David Luan, Dario Amodei, Ilya
  Sutskever, et~al. 2019.
\newblock Language models are unsupervised multitask learners.
\newblock volume~1, page~9.

\bibitem[{Raffel et~al.(2020)Raffel, Shazeer, Roberts, Lee, Narang, Matena,
  Zhou, Li, and Liu}]{Raffel2020ExploringTL}
Colin Raffel, Noam~M. Shazeer, Adam Roberts, Katherine Lee, Sharan Narang,
  Michael Matena, Yanqi Zhou, W.~Li, and Peter~J. Liu. 2020.
\newblock Exploring the limits of transfer learning with a unified text-to-text
  transformer.
\newblock \emph{J. Mach. Learn. Res.}, 21(140):1--67.

\bibitem[{Raina and Gales(2022)}]{raina-gales-2022-answer}
Vatsal Raina and Mark Gales. 2022.
\newblock \href {https://doi.org/10.18653/v1/2022.findings-acl.82} {Answer
  uncertainty and unanswerability in multiple-choice machine reading
  comprehension}.
\newblock In \emph{Findings of the Association for Computational Linguistics:
  ACL 2022}, pages 1020--1034, Dublin, Ireland. Association for Computational
  Linguistics.

\bibitem[{Rajpurkar et~al.(2018)Rajpurkar, Jia, and
  Liang}]{rajpurkar-etal-2018-know}
Pranav Rajpurkar, Robin Jia, and Percy Liang. 2018.
\newblock \href {https://doi.org/10.18653/v1/P18-2124} {Know what you don{'}t
  know: Unanswerable questions for {SQ}u{AD}}.
\newblock In \emph{Proceedings of the 56th Annual Meeting of the Association
  for Computational Linguistics (Volume 2: Short Papers)}, pages 784--789,
  Melbourne, Australia. Association for Computational Linguistics.

\bibitem[{Rajpurkar et~al.(2016)Rajpurkar, Zhang, Lopyrev, and
  Liang}]{rajpurkar-etal-2016-squad}
Pranav Rajpurkar, Jian Zhang, Konstantin Lopyrev, and Percy Liang. 2016.
\newblock \href {https://doi.org/10.18653/v1/D16-1264} {{SQ}u{AD}: 100,000+
  questions for machine comprehension of text}.
\newblock In \emph{Proceedings of the 2016 Conference on Empirical Methods in
  Natural Language Processing}, pages 2383--2392, Austin, Texas. Association
  for Computational Linguistics.

\bibitem[{Varanasi et~al.(2020)Varanasi, Amin, and
  Neumann}]{Varanasi2020CopyBERTAU}
Stalin Varanasi, Saadullah Amin, and G{\"u}nter Neumann. 2020.
\newblock Copybert: A unified approach to question generation with
  self-attention.
\newblock In \emph{Proceedings of the 2nd Workshop on Natural Language
  Processing for Conversational AI}, pages 25--31.

\bibitem[{Vu and Moschitti(2021)}]{vu-moschitti-2021-ava}
Thuy Vu and Alessandro Moschitti. 2021.
\newblock \href {https://doi.org/10.18653/v1/2021.naacl-main.412} {{AVA}: an
  automatic e{V}aluation approach for question answering systems}.
\newblock In \emph{Proceedings of the 2021 Conference of the North American
  Chapter of the Association for Computational Linguistics: Human Language
  Technologies}, pages 5223--5233, Online. Association for Computational
  Linguistics.

\bibitem[{Wang et~al.(2018)Wang, Lan, Nie, Waters, Grimaldi, and
  Baraniuk}]{wang2018answer}
Zichao Wang, Andrew~S. Lan, Weili Nie, Andrew~E. Waters, Phillip~J. Grimaldi,
  and Richard~G. Baraniuk. 2018.
\newblock \href {https://doi.org/10.1145/3231644.3231654} {Qg-net: A
  data-driven question generation model for educational content}.
\newblock In \emph{Proceedings of the Fifth Annual ACM Conference on Learning
  at Scale}, L@S '18, New York, NY, USA. Association for Computing Machinery.

\bibitem[{Yang and Liu(1999)}]{yang1999re}
Yiming Yang and Xin Liu. 1999.
\newblock A re-examination of text categorization methods.
\newblock In \emph{Proceedings of the 22nd annual international ACM SIGIR
  conference on Research and development in information retrieval}, pages
  42--49.

\bibitem[{Yang et~al.(2018)Yang, Qi, Zhang, Bengio, Cohen, Salakhutdinov, and
  Manning}]{Yang2018HotpotQAAD}
Zhilin Yang, Peng Qi, Saizheng Zhang, Yoshua Bengio, William~W. Cohen,
  R.~Salakhutdinov, and Christopher~D. Manning. 2018.
\newblock Hotpotqa: A dataset for diverse, explainable multi-hop question
  answering.
\newblock In \emph{Proceedings of the 2018 Conference on Empirical Methods in
  Natural Language Processing}.

\bibitem[{Yu et~al.(2020)Yu, Jiang, Dong, and Feng}]{Yu2020ReClorAR}
Weihao Yu, Zihang Jiang, Yanfei Dong, and Jiashi Feng. 2020.
\newblock Reclor: A reading comprehension dataset requiring logical reasoning.
\newblock \emph{International Conference on Learning Representations},
  abs/2002.04326.

\bibitem[{Zeng et~al.(2020)Zeng, Li, Li, Hu, and Hu}]{Zeng2020ASO}
Changchang Zeng, Shaobo Li, Qin Li, Jie Hu, and Jianjun Hu. 2020.
\newblock A survey on machine reading comprehension—tasks, evaluation metrics
  and benchmark datasets.
\newblock \emph{Applied Sciences}, 10(21):7640.

\bibitem[{Zhang et~al.(2021)Zhang, Yang, and Zhao}]{Zhang2021RetrospectiveRF}
Zhuosheng Zhang, Junjie Yang, and Hai Zhao. 2021.
\newblock Retrospective reader for machine reading comprehension.
\newblock In \emph{AAAI}.

\bibitem[{Zhao et~al.(2018)Zhao, Ni, Ding, and Ke}]{Zhao2018ParagraphlevelNQ}
Yao Zhao, Xiaochuan Ni, Yuanyuan Ding, and Qifa Ke. 2018.
\newblock Paragraph-level neural question generation with maxout pointer and
  gated self-attention networks.
\newblock In \emph{Proceedings of the 2018 conference on empirical methods in
  natural language processing}, pages 3901--3910.

\bibitem[{Zheng et~al.(2018)Zheng, Ma, and Huang}]{zheng-etal-2018-multi}
Renjie Zheng, Mingbo Ma, and Liang Huang. 2018.
\newblock \href {https://doi.org/10.18653/v1/D18-1357} {Multi-reference
  training with pseudo-references for neural translation and text generation}.
\newblock In \emph{Proceedings of the 2018 Conference on Empirical Methods in
  Natural Language Processing}, pages 3188--3197, Brussels, Belgium.
  Association for Computational Linguistics.

\end{thebibliography}
\bibliographystyle{acl_natbib}

\newpage
.
\newpage

\appendix
\appendixpage

\section{Assessment of high-entropy sequences}
\label{sec:app_assess}
In sequence-to-sequence generation tasks, one of the greatest challenges is regarding the scope of multiple possible output sequences for a given input sequence. Typically multiple reference samples will be provided for each input sequence. A model's prediction is compared (using a performance metric such as BLEU, METEOR or ROUGE) against each of the references in turn and the performance is defined using the best performance score. However, practical limitations (e.g. expense of collecting references) means that the number of references may not fully be representative of the large number of acceptable output sequences. Consequently, a model's prediction may achieve a low performance score by not being similar to any of the references albeit being an acceptable output sequence. Therefore, it is clear that there is expected to be some relationship between the achievable performance score with the number of reference samples available. This section aims to theoretically investigate how the number of reference samples impacts the achievable performance for sequence-to-sequence tasks.

\subsection{Log-likelihood framework}

A discriminative model, $\mathcal{M}$, defined by parameters $\boldsymbol{\theta}$, predicts the conditional representation based on its setting of $\boldsymbol{\theta}$:
\begin{equation}
   P_{\mathcal{M}}(y_{1:T} | x_{1:K}; \boldsymbol{\theta}). 
\end{equation}

A theoretical measure of the performance of a model (on some test sample) is given by the log-likelihood of the output from the trained model ($y_{1:T}\sim P_{\mathcal{M}}(y_{1:T}|x_{1:K};\boldsymbol{\theta})$) where the log-likelihood is using the true posterior. Therefore, the performance can be defined as:
\begin{equation}
    \log(P(y_{1:T}|x_{1:K}))
\end{equation}
In order to get the expected performance over the output sequence, the expectation is taken with respect to the conditional distribution learnt by the discriminative model:
\begin{equation}
     \underset{P_{\mathcal{M}}(y_{1:T}|x_{1:K};\boldsymbol{\theta})}{\mathbb{E}} \log(P(y_{1:T}|x_{1:K})).
\end{equation}
In an ideal training scenario,
\begin{equation}
P_{\mathcal{M}}(y_{1:T}|x_{1:K};\boldsymbol{\theta}) \rightarrow P(y_{1:T}|x_{1:K}).
\end{equation}
Therefore,
\begin{align}
   & \underset{P_{\mathcal{M}}(y_{1:T}|x_{1:K};\boldsymbol{\theta})}{\mathbb {E}} \log(P(y_{1:T}|x_{1:K})) \\
   & \rightarrow \underset{P(y_{1:T}|x_{1:K})}{\mathbb{E}} \log(P(y_{1:T}|x_{1:K})) \\
  &  = -H(Y_{1:T}|X_{1:K}=x_{1:K}),
\end{align}
where $H(\cdot)$ denotes the entropy of a probability distribution. So far, we have only taken expectations over the output sequence when conditioned on a particular input sequence. However, we are interested in finding the expected performance not only over the output sequence but also by considering all possible input sequences that we may be conditioned by. Thus, an expectation must be taken over the input distribution too to define the overall expected performance over the test dataset. Performance: 
\begin{align}
         &= \underset{P(x_{1:K})}{\mathbb{E}}\left[\underset{P(y_{1:T}|x_{1:K})}{\mathbb{E}} \log(P(y_{1:T}|x_{1:K}))\right] \\
        &= -\underset{P(x_{1:K})}{\mathbb{E}}\left[H(Y_{1:T}|X_{1:K}=x_{1:K})\right] \\
        &= -H(Y_{1:T}|X_{1:K})
\end{align}

The above expression is the conditional entropy of the output sequence given the input sequence. Hence, greater the uncertainty in the conditional distribution, lower the expected performance on the test set. It is noteworthy that from the above expression the maximum performance is a value of 0 (for a sharp distribution) and the minimum performance is given by $-\log M$ (for a uniform distribution) where $M$ denotes the total number of possible output sequences. The uncertainty in the conditional distribution is dependent on the task. Hence, it is natural to expect that high performance cannot be achieved for question generation tasks when there is only one reference sample provided for a given input in the test dataset.

\subsection{Multiple draws}

In practical sequence-to-sequence tasks, multiple references (labelled output sequences) are given for each input sequence in order to try and capture the number of acceptable output sequences for tasks that have high uncertainty (e.g. machine translation, summarisation, question generation). At evaluation time, typically the reference that allows the best performance to be achieved is selected. It is interesting to consider, in the theoretical sense, how many output sequence samples are required to be drawn from the posterior (conditional) distribution in order to achieve a threshold performance. 

Let us denote the number of drawn output sequences per input sequence as $J$. Therefore, we have the pair $\left(x_{1:K}, \left\{y_{1:T}^{(j)}\right\}_{j=1}^J\right)$, where $x_{1:K} \sim P(x_{1:K})$ and $y_{1:T}^{(j)}\sim P(y_{1:T}|x_{1:K})$ for $j\in[1,J]$. We assume that each of the output sequences are drawn independently of each other from the conditional distribution. So, we are basically interested in the impact of $J$ on the expected performance. The expected multi-draw performance is then given as $\underset{P(x_{1:K})}{\mathbb{E}}$ of:
\begin{multline}
         \underset{P(y_{1:T}^{(1)},y_{1:T}^{(2)},\hdots,y_{1:T}^{(J)}|x_{1:K})}{\mathbb{E}}
         \\
         \left[\max_{j\in[1,J]}\left\{ \log(P(y_{1:T}^{(j)}|x_{1:K}))\right\}\right]
\end{multline}

However, it is not possible to analytically simplify the above equation. Therefore, a different mathematical framework will be considered to deduce the impact on performance of multiple draws of the reference sample from the true posterior distribution. The alternative frameworks are considered in the below sections.

\subsection{Maximum likelihood output with exact match framework}

Let us define the output prediction from a trained model, $y_{1:T}^*$, as the most likely output.
\begin{equation}
y_{1:T}^* = \text{arg} \max_{y_{1:T}} \left\{P_{\mathcal{M}}(y_{1:T}|x_{1:K};\boldsymbol{\theta})\right\}    
\end{equation}
As before, we consider the ideal case where
\begin{equation}
P_{\mathcal{M}}(y_{1:T}|x_{1:K};\boldsymbol{\theta}) \rightarrow P(y_{1:T}|x_{1:K}).
\end{equation}
Therefore $y_{1:T}^*$ is the modal output from the true posterior distribution,
\begin{equation}
y_{1:T}^* = \text{arg} \max_{y_{1:T}} \left\{P(y_{1:T}|x_{1:K})\right\}    
\end{equation}
Let $J$ reference samples be sampled from the true posterior distribution,
\begin{equation}
    \hat{y}_{1:\hat{T}}^{(j)}\sim P(y_{1:T}|x_{1:K}) \hspace{10pt} \text{for } j\in[1,J]
\end{equation}
Then, let us define the performance for a given prediction as:
\begin{equation}
    \max_{j\in[1,J]}\left\{F\left[\hat{y}_{1:\hat{T}}^{(j)},y_{1:T}^*\right]\right\}
\end{equation}
where $F(\cdot)$ represents the exact match performance metric i.e. the predicted sequence has to exactly match one of the reference sequences.
Thus, the expected multi-draw performance is given by:
\tiny
\begin{align}
\label{eq:frame1}
         &= \underset{P(x_{1:K})}{\mathbb{E}}\left[\underset{P(\hat{y}_{1:\hat{T}}^{(1)},\hat{y}_{1:\hat{T}}^{(2)},\hdots,\hat{y}_{1:\hat{T}}^{(J)}|x_{1:K})}{\mathbb{E}}\left[\max_{j\in[1,J]}\left\{  F\left[\hat{y}_{1:\hat{T}}^{(j)},y_{1:T}^*\right]  \right\}\right]\right] \\
        &= 
        \underset{P(x_{1:K})}{\mathbb{E}}\left[ P\left( \left( \hat{y}_{1:\hat{T}}^{(1)} = y_{1:T}^*\right) \cup \left(\hat{y}_{1:\hat{T}}^{(2)}=y_{1:T}^*\right) \cup \hdots   \right) \right] \\
        &=
        \underset{P(x_{1:K})}{\mathbb{E}}\left[ 1 - P\left( \left( \hat{y}_{1:\hat{T}}^{(1)}\neq y_{1:T}^*\right) \cap \left( \hat{y}_{1:\hat{T}}^{(2)}\neq y_{1:T}^*\right) \cap \hdots  \right) \right] \\
        &=
        \underset{P(x_{1:K})}{\mathbb{E}}\left[ 1 - P \left( \hat{y}_{1:\hat{T}}^{(1)}\neq y_{1:T}^*\right) P\left( \hat{y}_{1:\hat{T}}^{(2)}\neq y_{1:T}^*\right) \hdots  \right] \\
        &=
        \underset{P(x_{1:K})}{\mathbb{E}}\left[ 1 - \left(1-P\left(y_{1:T}^*|x_{1:K}\right)\right)^J   \right] \\
        & \approx 
        \underset{P(x_{1:K})}{\mathbb{E}}\left[ 1 - \left(1-JP\left(y_{1:T}^*|x_{1:K}\right)\right)   \right] \hspace{2pt} \text{assuming }  P\left(y_{1:T}^*|x_{1:K}\right) << 1  \\
        &=
        J\underset{P(x_{1:K})}{\mathbb{E}}\left[P\left(y_{1:T}^*|x_{1:K}\right)\right]\\ 
        &= 
        J \times \text{constant}
\end{align}
\normalsize
Therefore, for the situation the mode of the posterior distribution is small (this is likely to be the case for high entropy probability distributions such as question generation), we can expect that the performance increases linearly with number of reference samples drawn from the true posterior distribution. This is an expected result as we are essentially saying that each reference sample is independently drawn of all other reference samples.

\subsection{Maximum likelihood output with fraction overlap score framework}

As before, the modal output from the true posterior distribution and the $J$ reference samples from the true distribution have the same definitions:
\begin{equation}
y_{1:T}^* = \text{arg} \max_{y_{1:T}} \left\{P(y_{1:T}|x_{1:K})\right\}    
\end{equation}

\begin{equation}
    \hat{y}_{1:\hat{T}}^{(j)}\sim P(y_{1:T}|x_{1:K}) \hspace{10pt} \text{for } j\in[1,J]
\end{equation}

\noindent Here, the precision of the overlap between the prediction and reference sample will be used (recall divides by the length of the reference sample). Multi-draw metric:

\begin{equation}
 = \max_{j\in[1,J]}\left\{F\left[\hat{y}_{1:\hat{T}}^{(j)},y_{1:T}^*\right]\right\}
\end{equation}
where
\begin{equation}
F[\mathbf{a},\mathbf{b}] = \frac{\mathbf{a} \cap \mathbf{b}}{|\mathbf{b}|} 
\end{equation}

The focus will be on the overlap of unigrams but the argument can easily be extended for larger n-grams. Note, we are not really concerned about different lengths of sequences (i.e $T$ and $\hat{T}$) because theoretically we can simply consider only sequences of all the same fixed length where an empty character can be thought of as a unigram in its own right; in this way the argument demonstrated here applies to sequences of all lengths.

Let us denote the prediction with $\alpha$ unigrams changed at some given positions across the sequence, where the specific positions being changed is given by a configuration function $c(\cdot)$ as $y_{1:T\setminus c(\alpha) }^*$. Then it will be useful to note that:
\begin{equation}
    F\left[y_{1:T}^*,y_{1:T\setminus c(\alpha)}^*\right] = \frac{T-\alpha}{T}
\end{equation}

Therefore, the expected multi-draw performance is given by (some steps use results directly from the working in the previous section):
\tiny
\begin{align}
\label{eq:frame2}
 &= \underset{P(x_{1:K})}{\mathbb{E}}\left[\underset{P(\hat{y}_{1:\hat{T}}^{(1)},\hat{y}_{1:\hat{T}}^{(2)},\hdots,\hat{y}_{1:\hat{T}}^{(J)}|x_{1:K})}{\mathbb{E}}\left[\max_{j\in[1,J]}\left\{  F\left[\hat{y}_{1:\hat{T}}^{(j)},y_{1:T}^*\right]  \right\}\right]\right] \\
        &= 
        \underset{P(x_{1:K})}{\mathbb{E}}\left[ 
        \sum_{\alpha=0}^{T} \frac{T-\alpha}{T} \sum_{c\in \mathcal{C}} P\left( \left( \hat{y}_{1:\hat{T}}^{(1)} = y_{1:T\setminus c(\alpha)}^*\right) \cup \hdots   \right) 
        \right] \\
        &\approx
        \underset{P(x_{1:K})}{\mathbb{E}}\left[ 
        \sum_{\alpha=0}^{T} \frac{T-\alpha}{T} \sum_{c\in \mathcal{C}} J P\left(y_{1:T\setminus c(\alpha)}^* | x_{1:K} \right) 
        \right] \\
        &= 
        J\underset{P(x_{1:K})}{\mathbb{E}}\left[ 
        \sum_{\alpha=0}^{T} \frac{T-\alpha}{T} \sum_{c\in \mathcal{C}}  P\left(y_{1:T\setminus c(\alpha)}^* | x_{1:K} \right) 
        \right] \\
        &= 
        J \times \text{constant}
\end{align}
\normalsize

The maximum likelihood output with an exact match framework and the maximum likelihood output with a fraction overlap score framework both indicated that there is a linear relationship between the performance and the number of references. Note, the theoretical arguments have assumed that there are a large number of possible output sequences and hence the probability of a prediction matching a reference sample (where both are sampled from the same true posterior distribution) is very small. $J$ references are sampled and it is assumed that each of these references is unique because we are only considering a small number of references and a large number of possible output sequences (i.e. high entropy posterior). This leads to a linear relationship between the multi-draw performance and the number of references which is \textit{only valid while the number of references is small}. The behaviour for a large number of reference samples can be seen to grow non-linearly prior to the assumption being applied. Note, the linear relationship for independently drawn references is an obvious result intuitively as each independence allows the probability of matching any one reference to distinctly be added up together. 

\section{Training details}

This section specifies the training details for all the systems trained and discussed in this work. All hyperparamter tuning was performed on the validation (development) set.

\subsection{Multiple-choice machine reading comprehension}
\label{sec:app_trn_mcmrc}
An ensemble of 3 models were trained using the ELECTRA-large pretrained language model. Each model in the ensemble has 340M parameters. Optimal hyperparameter values were motivated from \citet{raina-gales-2022-answer} who performed a grid search to identify the best settings for performance on the RACE dataset for multiple-choice machine reading comprehension. In particular, each ensemble member was trained for 2 epochs at a learning rate of 2e-6 with a batch size of 4 and inputs truncated to 512 tokens. Cross-entropy loss was used at training time with models built using NVIDIA V100 graphical processing units with training time under 8 hours per model.

\subsection{Question complexity}
\label{sec:app_trn_qc}
An ensemble of 3 models were trained using the ELECTRA-large pretrained language model. Each model in the ensemble has 340M parameters. The selection of hyperparameter values was achieved by performing a grid search on the learning rate $\in \{2e-7, 2e-6, 2e-5\}$ and the batch size $\in \{2,4\}$. In particular, each ensemble member was trained for 2 epochs at a learning rate of 2e-6 with a batch size of 4 and inputs truncated to 512 tokens. Cross-entropy loss was used at training time with models built using NVIDIA V100 graphical processing units with training time under 7 hours per model.

\subsection{Question generation}
\label{sec:app_trn_qg}
A single question generation model was trained using the T5-base pretrained language model. The model has 220M parameters. The selection of hyperparameter values was achieved by performing a grid search on the learning rate $\in \{2e-7, 2e-6, 2e-5\}$ and the batch size $\in \{2,4\}$. In particular, the model was trained for 3 epochs at a learning rate of 2e-6 with a batch size of 4. The inputs were truncated to 512 tokens. Cross-entropy loss was used at training time with the model built using NVIDIA V100 graphical processing units with training time under 12 hours. At decoding time, a deterministic beam size of 4 was used with a repetition penalty of 2.5, a length penalty of 1.0, maximum sequence length of 80 tokens and early stopping (reaching end token) permitted.

\section{Example question generated}
\label{sec:app_example}
Here, the example question generated demonstrates a failure mode of the proposed multiple-choice question generation approach as two of the distractor options are indeed valid answer options too. 

\begin{tcolorbox}
\begin{small}
\begin{tabular}{@{}p{6.7cm}}
\textbf{Context:} \\
When she graduates from Columbia University next year with a master's degree in Public health, Eric Wheeler is hoping to get a job in international reproductive health. The 26-year-old post-graduate has always wanted to work in public service. But public service doesn't pay much, and her two-year program at Columbia costs about \$50,000 a year with living expenses. She has a scholarship from Columbia that covers just \$4,000 a year and has taken out loans to pay for the rest. She worries that she will spend years paying back her student loans and not have money left over to put away in an IRA. \\
... \\
Typically, it is projected that a borrower who performs public service under this program will repay only about one-fourth to one-half as much money as a borrower who does not\", he said. He also pointed out that public service is broadly defined and includes any government and nonprofit organization job. \\ \\

\textbf{Question:}  \\
Which of the following is TRUE according to the passage?

\end{tabular}

\textbf{Options:}
\newline
\begin{tabular}{@{}lp{6cm}}
(A)  & Wheeler wants to get a job in international reproductive health. \\
(B) & Wheeler's two-year program at Columbia costs about \$50,000 a year. \\
(C) & Wheeler has taken out loans to pay for the rest. \\
(D) & Wheeler will spend years paying back her student loans.
\end{tabular}
\end{small}
\end{tcolorbox}

\newpage

\section{Additional results}

This section simply aims to present the standard deviations of the single seed results (insufficient space in the main text) across the 3 models in an ensemble for both the multiple-choice machine reading comprehension systems and the question complexity systems. Additionally, the results on the Dev set are given for the single seed models too for multiple-choice machine reading comprehension.

\subsection{Multiple-choice machine reading comprehension}
\label{sec:app_additional_mcmrc}
\begin{table}[htbp!]
\centering
\begin{small}
    \begin{tabular}{l|rrrr}
    \toprule
System & Dev & Evl \\
\midrule
M & $88.70_{\pm 0.18}$ & $88.30_{\pm0.15}$ \\
H & $85.59_{\pm 0.18}$ & $84.11_{\pm0.15}$ \\
C & $81.98_{\pm 0.57}$ & $81.07_{\pm0.46}$ \\
All & $85.93_{\pm 0.21}$ & $84.80_{\pm0.12}$ \\
   \bottomrule
    \end{tabular}
    \end{small}
\caption{Single seed results for multiple-choice machine reading comprehension on Dev and Evl of RACE++ assessed using accuracy (\%).}
    \label{app_tab:res_mrc}
\end{table}

\subsection{Question complexity}
\label{sec:app_additional_qc}
\begin{table}[htbp!]
\centering
\begin{small}
    \begin{tabular}{l|rrrr}
    \toprule
System & Dev & Evl \\
\midrule
Accuracy & $86.00_{\pm 0.36}$ & $86.19_{\pm 0.50}$ \\
Macro F1 & $83.92_{\pm 0.37}$ & $84.66_{\pm 0.53}$ \\
   \bottomrule
    \end{tabular}
    \end{small}
\caption{Single seed results for question complexity on Dev and Evl of RACE++.}
    \label{app_tab:res_qc}
\end{table}

\end{document}